# A Step Forward in Studying the Compact Genetic Algorithm

**R. Rastegar**  rrastegar@ieee.org
Department of Mathematics, Southern Illinois University, Carbondale, IL 62901, US

**A. Hariri**  hariri@ieee.org
Iran Telecommunication Research Center, North Karegar St., Tehran, Iran

**Abstract**
The compact Genetic Algorithm (cGA) is an Estimation of Distribution Algorithm that generates offspring population according to the estimated probabilistic model of the parent population instead of using traditional recombination and mutation operators. The cGA only needs a small amount of memory; therefore, it may be quite useful in memory-constrained applications. This paper introduces a theoretical framework for studying the cGA from the convergence point of view in which, we model the cGA by a Markov process and approximate its behavior using an Ordinary Differential Equation (ODE). Then, we prove that the corresponding ODE converges to local optima and stays there. Consequently, we conclude that the cGA will converge to the local optima of the function to be optimized.

## 1 Introduction

One of the most famous optimization procedures for combinatorial optimization is the Genetic Algorithm (GA). By maintaining a population of solutions, the GA can be viewed as implicitly modeling of the solutions seen in the search process. In the standard GA, new solutions are generated by applying randomized recombination operators on two or more high-quality individuals of the current population (Goldberg, 1989). These recombination operators, such as one-point, two-point or uniform crossover, randomly select non-overlapping subsets of two "parent" solutions to form "children" solutions. By using a crossover operator that preserves groups of parameters from parents to children, the GA attempts to capture dependencies between the parameters implicitly.

The poor behavior of genetic algorithms in some problems, sometimes attributed to designed operators, has led to the development of other types of algorithms. The Probabilistic Model Building Genetic Algorithms (PMBGAs) or Estimation of Distribution Algorithms (EDAs) are a class of algorithms which has been developed recently to preserve the building blocks (Larranaga and Lozano, 2001). The principal concept in this new technique is to prevent the disruption of partial solutions contained in a solution by building a probabilistic model. The EDAs are classified into three classes based on the interdependencies between the variables of solutions (Larranaga and Lozano, 2001) (Pelikan et al., 1999); the no dependencies model, the bivariate dependencies model, and the multiple dependencies model. To name just a few, instances of EDA algorithms include the Population-based Incremental Learning (PBIL) (Baluja, 1994) (Baluja and Caruana, 1995), the Bit-based Simulated Crossover (BSC) (Syswerda, 1992), the Univariate Marginal Distribution Algorithm (UMDA) (Muhlenbein, 1998), the compact Genetic Algorithm, (cGA) (Harik et al., 1999) and the Learning Automata based Estimation of Distribution Algorithm (LAEDA) (Rastegar and Meybodi, 2005a) for the no dependencies model, Mutual Information Maximization for Input Clustering (MIMIC) (De Bonet et al., 1996) and Combining Optimizer with Mutual Information Trees (COMIT) (Baluja and Davies, 1997) for the bivariate dependencies model, and finally, the Factorized Distribution Algorithm (FDA) (Muhlenbein and Mahnig, 1999b) and the Bayesian Optimization Algorithm (BOA) (Pelikan et al., 1999) for the multiple dependencies model.

Some researchers have studied the working mechanism of EDAs. The behavior of the UMDA and the PBIL has been studied in (Muhlenbein, 1998) (Gonzalez et al., 2000) (Hohfeld and Rudolph, 1997) (Zhang, 2004a) (Rastegar and Meybodi, 2005b) (Neil and Rowe, 2005). Both (Zhang, 2004b) and (Muhlenbein and Mahnig, 1999a) discuss the convergence of the FDA for separable additively decomposable functions. In (Zhang and Muhlenbein, 2004), Zhang and Mhlenbein have proven that a class of EDAs with an infinite population size globally converges. (Rastegar and Meybodi, 2005c) carried out a study on the time complexity of EDAs with an infinite population size.

Although all algorithms for the no dependencies model have low efficiency in solving difficult problems, it is still important to study them due to their simplicity in terms of memory usage and computational complexity and with respect to the fact that the computational complexity of the bivariate dependencies model and the multiple dependencies model is high. One of the simplest algorithms of the no dependencies model is the compact Genetic Algorithm (cGA). This algorithm initializes a Probability Vector (PV), where each component of the PV follows a Bernoulli distribution with the parameter of 0.5, and then two solutions are randomly generated by using this PV. The generated solutions are ranked based on their fitness values. Then, the PV is updated based on these solutions. This process of adaptation continues until the PV converges. The cGA represents the population as a PV over a set of solutions and operationally mimics the order-one behavior of the Simple GA (SGA) with the uniform crossover. When confronted with easy problems (e.g., continuous-unimodal problems involving lower order BBs), the cGA achieves the performance of the SGA (with the uniform crossover) in terms of the number of fitness evaluations (Harik et al., 1999).

By now, some variations of cGA have been introduced and some of them have been utilized in real world applications (Ahn and Ramakrishna, 2003) (Gallagher et al., 2004) (Baraglia et al., 2001a) (Sastry and Goldberg, 2000) (Baraglia et al., 2001b), but the behavior of the cGA has not been studied in details. To the best knowledge of the authors, the only papers in which the analytical analysis of the cGA has been done are (Ahn and Ramakrishna, 2003) and (Droste, 2005). In (Ahn and Ramakrishna, 2003), it has been proven that the elitism-based cGA is equal to the Evolution Strategy (ES). In (Droste, 2005), Droste has presented the first rigorous runtime analysis of the cGA for linear pseudo-boolean functions. He has shown that not all linear functions have the

*Parameters:* $\alpha$ is the learning step, $n$ is the solution length
*Step 1.* Set $k$ to $0$, and initialize the probability vector
    For $i := 1$ to $n$ do $p_i(k) := 0.5$;
*Step 2.* Generate two solutions from the probability vector
    $a(k) := \text{generate}(p(k))$; $b(k) := \text{generate}(p(k))$;
*Step 3.* Let them compete
    $w(k), l(k) := \text{compete}(a(k), b(k))$;
    where $w(k)$ and $l(k)$ are winner and loser solutions respectively.
    (if both $a(k)$ and $b(k)$ have the same fitness value then $a(k)$ is selected as $w(k)$)
*Step 4.* Update the probability vector
    For $i := 1$ to $n$ do
        If $w_i(k) \neq l_i(k)$ then
            If $w_i(k) == 1$ then $p_i(k+1) := p_i(k) + \alpha$;
            Else $p_i(k+1) := p_i(k) - \alpha$;
*Step 5.* Check if the probability vector has converged.
    Go to Step 2, if it is not satisfied.

Figure 1: Pseudocode of the cGA

same asymptotical runtime. In this paper, we study the cGA as a recursive stochastic algorithm. We model the cGA as a Markov process and approximate it by an ODE where its learning step is small. Then, we study the behavior of the obtained ODE and determine its convergence and stability properties.

This work is organized as follows. Section 2 describes the cGA precisely. In Section 3, a formulation of the cGA and some required definitions and lemmas are stated. In Section 4, the analysis of the cGA as a Markov process is done in two stages. In the first stage, we derive an ODE whose solution approximates the asymptotic behavior of the cGA. Then in the second stage, we prove that the corresponding ODE and therefore, the cGA surely converge to the local optima of the function to be optimized and stays at them. Finally, Section 5 concludes the paper.

## 2 The Compact Genetic Algorithm

At each iteration $k$, the cGA manages its population as a PV, $p(k) = (p_1(k), ..., p_n(k))$, where $n$ is the number of genes, thereby it mimics the order-one behavior of the SGA with the uniform crossover (Harik et al., 1999). The value of $p_i(k) \in [0,1]$, $i = 1, ..., n$, measures the proportion of the allele "1" in the $i$th locus of the simulated population. Figure 1 describes the pseudocode of the cGA.

For $i = 1, ..., n$, $p_i(0)$ is initialized with 0.5 to represent a randomly generated population. In each generation (i.e. iteration), two competing solutions are generated on the basis of the current PV and then the PV is updated to favor the better solution (i.e. winner). The probability $p_i(k)$ is increased (decreased) by the learning step, $\alpha$, when the $i$th locus of the winner has an allele of "1" (resp. "0") and the $i$th locus of the loser has an allele of "0" (resp. "1"). If both the winner and the loser have the same allele in the $i$th locus, then the probability $p_i(k)$ remains the same. This scheme is equivalent to the (steady-state) pair-wise tournament selection (Harik et al., 1999). The cGA terminates when all the probabilities converge to zero or one.

## 3  Problem Formulation

Let $y = (y_1, ..., y_n)$ denote a solution where $y_i$ belongs to $\{0, 1\}$ and consider that $g : \Omega \to \Re$ is an injective pseudo-boolean function to be maximized, where $\Omega = \{0, 1\}^n$. The goal is to maximize $g$ using the cGA. At the $k$th iteration of the optimization process, two solutions $w(k)$ and $l(k)$ are generated on the basis of $p(k)$ where $g(w(k)) \geq g(l(k))$, and then, the PV is updated as follows,

$$p_i(0) = 0.5 \, , \, 1 \leq i \leq n$$

$$p(k+1) = p(k) + \alpha(w(k) - l(k)) \quad (1)$$

To prevent $p_i$s from getting smaller than 0 or larger than 1, we let $\alpha$ be equal to $1/(2N)$, where $N$ is a positive integer number. In the remainder of this section, we introduce our definitions and derive some results that will be used later for the analysis of the cGA.

**Definition 1.** A solution $y$ is called a local maximum of the function $g$, if and only if, for each solution $z$, whose hamming distance to the solution $y$ is one, i.e. $d_H(y, z) = 1$, we have $g(y) \geq g(z)$. A local maximum is called strict if the inequality is strict.

**Definition 2.** The configuration space of the cGA is $K = [0, 1]^n$ where $p(k) \in K$ for each $k$. Also $K^* = \{0, 1\}^n$ ($K^*$ is equivalent to $\Omega$) is called the corner (the deterministic subspace) of $K$ and $K - K^*$ is called the non-deterministic subspace of $K$.

**Definition 3.** $p(k)$ is called a deterministic configuration if $p_i(k) = 0$ or $1$ for every $i = 1, ..., n$, i.e. $p(k) \in K^*$, in the other cases, $p(k)$ is called a non-deterministic configuration, i.e. $p(k) \in K - K^*$.

**Lemma 1.** Let $Pr(w(k) = y|p(k))$ be the probability of obtaining $y$ as the winner solution of the $k$th iteration. Then

$$Pr(w(k) = y|p(k)) = Pr(y|p(k))\{\sum_{g(z) < g(y)} Pr(z|p(k)) + \sum_{g(z) \leq g(y)} Pr(z|p(k))\} \quad (2)$$

where $Pr(y|p(k))$ denotes the probability of sampling the solution $y$.

**Proof.** At each iteration, two solutions are sampled from $p(k)$. The probability that the first sampled solution is equal to $y$ and be the winner solution is $Pr(y|p(k))Pr(\text{all } z|p(k), g(y) \geq g(z))$, and the probability that the second sampled solution is the winner solution and be equal to $y$ is $Pr(\text{all } z|p(k), g(z) < g(y))Pr(y|p(k))$. Therefore, $Pr(w(k) = y|p(k))$ is equal to the sum of these probabilities. Hence the proof. ◇

**Lemma 2.** Let $Pr(l(k) = y|p(k))$ be the probability of obtaining $y$ as the loser solution of the $k$th iteration. Then

$$Pr(l(k) = y|p(k)) = Pr(y|p(k))\{\sum_{g(z)>g(y)} Pr(z|p(k)) + \sum_{g(z)\geq g(y)} Pr(z|p(k))\} \quad (3)$$

where $Pr(y|p(k))$ denotes the probability of sampling the solution $y$. Proof is similar to the proof of Lemma 1.

**Lemma 3.** Assume that $p_m$ and $y_m$ are the $m$th positions of $p$ and $y$ respectively. Then equations (4)-(9) are true for $Pr(y|p)$,

$$Pr(y|p) = \prod_{i=1}^{n} p_i^{y_i}(1-p_i)^{1-y_i} \quad (4)$$

$$Pr(y|p) = \begin{cases} 0 & \text{for } p \in K^*, \ p \neq y \\ 1 & \text{for } p \in K^*, \ p = y \end{cases} \quad (5)$$

$$\left.\frac{\partial Pr(y|p)}{\partial p_m}\right|_y = \begin{cases} 1 & \text{if } y_m = 1 \\ -1 & \text{if } y_m = 0 \end{cases} \quad (6)$$

$$\left.\frac{\partial Pr(z|p)}{\partial p_m}\right|_y = 0 \quad \text{for all } z \text{ whose } d_H(z,y) \geq 2 \quad (7)$$

$$\left.\frac{\partial Pr(z|p)}{\partial p_m}\right|_y = \begin{cases} 1 & \text{if } d_H(z,y) = 1 \text{ and } z_m = 1, \ y_m = 0 \\ -1 & \text{if } d_H(z,y) = 1 \text{ and } z_m = 0, \ y_m = 1 \end{cases} \quad (8)$$

$$\left.\frac{\partial Pr(z|p)}{\partial p_m}\right|_y = 0 \quad \text{if } d_H(z,y) = 1 \text{ and } y_m = z_m \quad (9)$$

where $y \in \Omega$ and $Pr(y|p)$ denotes the probability of sampling the solution $y$.
**Proof.** Equation (4) is trivial by the fact that all $y_i$s are independent. The other results can be easily obtained by (4) (Gonzalez et al., 2000). ⋄

## 4 Analysis of the Compact Genetic Algorithm

Under the algorithm specified by (1), $\{p(k), k \geq 0\}$ is a Markov process. The analysis of this process is done in two stages. In the first stage, we derive an ODE whose solutions approximate the asymptotic behavior of $p(k)$ for a sufficiently small learning step $\alpha$ (i.e. $N$ tends to infinity) used in (1). In the second stage, we characterize the solutions of the ODE and thus, we obtain the long-term behavior of $p(k)$.
The algorithm given by (1) can be represented as

$$p(k+1) = p(k) + \alpha G(p(k), w(k), l(k)), \text{ where } G(p(k), w(k), l(k)) = w(k) - l(k) \quad (10)$$

$w(k)$ and $l(k)$ denote the winner and the loser solutions respectively and $\alpha$ is the learning step. Now, define

$$\Delta p(k) = E\{p(k+1)|p(k)\} - p(k) \quad (11)$$

where $E\{.\}$ is the mathematical expectation operator. Since $\{p(k); k \geq 0\}$ is Markovian and $w(k)$ and $l(k)$ only depend on $p(k)$ not on $k$, then $\Delta p(k)$ can be given as follows.

$$\Delta p(k) = \alpha f(p(k)) \tag{12}$$

where $f : K \to K$ and

$$f(p) = E\{G(p(k), w(k), l(k)) | p(k) = p\} = E\{w(k) - l(k) | p(k) = p\} \tag{13}$$

The function $f(p)$ can be rewritten as follows,

$$\begin{aligned} f(p) &= E\{w(k)|p(k)\} - E\{l(k)|p(k)\} \\ &= \sum_y y Pr(w(k) = y|p(k)) - \sum_y y Pr(l(k) = y|p(k)) \\ &= \sum_y y (Pr(w(k) = y|p(k)) - Pr(l(k) = y|p(k))) \end{aligned} \tag{14}$$

By Lemma 1 and Lemma 2, and some simplification, we have

$$f(p) = 2 \sum_y y Pr(y|p(k)) \{ \sum_{g(z)<g(y)} Pr(z|p(k)) - \sum_{g(z)>g(y)} Pr(z|p(k)) \} \tag{15}$$

Now, define a sequence of continuous-time interpolations of (10) denoted by $p^\alpha(t)$ and called an interpolated process, whose components are defined by

$$p_i^\alpha(t) = p_i(k), \ 1 \le i \le n, \ t \in [k\alpha, (k+1)\alpha) \tag{16}$$

The interpolated process $\{p^\alpha(t)\}_{t \ge 0}$ is a sequence of random variables that takes values in $D^n$, the space of all right continuous functions with left hand limits defined over $[0, \infty)$ and $p^\alpha$ takes values in $K$ which is a bounded subset of $\Re^n$. The objective is to study the limit behavior of the sequence $\{p^\alpha(t)\}_{t \ge 0}$ as $\alpha$ (resp. N) tends to zero (resp. infinity), which will be a good approximation of the asymptotic behavior of (16). When $\alpha$ tends to zero, (12) can be written as the following ODE

$$\frac{dp}{dt} = f(p) \tag{17}$$

We are interested in characterizing the long-term behavior of $p(k)$ and hence the asymptotic behavior of the ODE (17). Now, we show that the sequence of interpolated processes $\{p^\alpha(.)\}$ weakly converges to the solution of the ODE (17) with the initial configuration $p(0)$. This implies that asymptotic behavior of $p(k)$ can be obtained from the solution of the ODE (17).

**Theorem 1.** Consider the sequence of interpolated processes $\{p^\alpha(t)\}$. Let $X_0 = p^\alpha(0) = p(0)$. When $\alpha \to 0$, the sequence weakly converges to $X(.)$ which is the solution of the ODE,

$$\frac{dX}{dt} = f(X), X(0) = X_0 \tag{18}$$

**Proof.** The theorem is a particular case of a general result to weak convergence theorem ((Kushner, 1984), Theorem 3.2). We note the following about the cGA given by (1),

1. $\{p(k), (w(k-1), l(k-1)), k \ge 1\}$ is a Markov process.

2. $(w(k), l(k))$ takes values in a compact metric space.

A Step Forward in Studying the Compact Genetic Algorithm

3. The function $G(.,.,.)$ is bounded, continuous, and independent of $\alpha$.

4. For a specific configuration, $p(k) = p$, $\{(w(k), l(k)), k \geq 0\}$ is an independent identically distributed (i.i.d.) sequence. Let $M^p$ be the distribution of the process $\{(w(k), l(k)), k \geq 0\}$.

5. The ODE (18) has a unique solution for each initial condition.

Hence, by using the weak convergence theorem ((Kushner, 1984), Theorem 3.2), when $\alpha \to 0$, the sequence $\{p^\alpha(.)\}$ weakly converges to the solution of the ODE

$$\frac{dX}{dt} = \hat{G}(X), \ X(0) = X_0$$

where $\hat{G}(p) = E^p G(p(k), w(k), l(k))$ and $E^p$ denotes the expectation with respect to the invariant measure $M^p$. Since for $p(k) = p$, $(w(k), l(k))$ is an i.i.d. sequence whose distribution only depends on $p$ and the function $g$, we have

$$\hat{G}(p) = E\{G(p(k), w(k), l(k))|p(k) = p\} = f(p) \qquad (19)$$

Hence the theorem. $\diamond$

Theorem 1 enables us to understand the long-term behavior of $p(k)$. The weak convergence in this theorem implies that when $\alpha$ tends to zero, the trajectory of $p^\alpha(t)$ will closely follow the solution of the ODE with a high probability at any finite interval. As the length of time interval increases and $\alpha$ tends to zero, the trajectory of the ODE spends most of the time required by the optimization process in a small neighborhood of $p^0$, the solution of the ODE. Thus, $p^\alpha(.)$ will eventually (with a high probability) spend all of its time in a small neighborhood of $p^0$ as well. As $\alpha$ tends to zero, the cGA follows the trajectory of the ODE in a time interval, which tends to infinity. The above point is summarized in the following Lemma.

**Lemma 4.** For a large $k$ and a small enough value of $\alpha$, the asymptotic behavior of $p(k)$ can be approximated by the solution of the ODE (18) with the same initial configuration.
**Proof.** Let $X(.)$ be the solution of the ODE (18) with the initial condition of $X(0) = X_0$ which is sufficiently close to an asymptotically stable configuration of the ODE, say $p^0 \in K$. For any $Y(t) \in K, t \geq 0$ and any positive $T < \infty$, define

$$h_T(Y) = \sup_{0 \leq t \leq T} \|Y(t) - X(t)\| \qquad (20)$$

Function $h_T(.)$ is continuous on $K$. Theorem 1 states that $E\{h_T(p^\alpha)\}$ tends to $E\{h_T(X)\} = 0$ as $\alpha \to 0$, the limit is zero since the value of $h_T(X)$ on the trajectories of limit process is zero with probability one. Thus, the $sup$ of the distance between the original sequence $p(t)$ and $X(t)$ goes to zero in probability as $k$ tends to infinity. With the particular initial condition used, let $p^0$ be the stationary configuration to which the solution of the ODE converges. Using this and the nature of interpolation, given in (16), it is implied that for the given initial configuration, any $\epsilon > 0$, and the integers $K_1, K_2, 0 < K_1 < K_2 < \infty$, there exists an $\alpha_0$ such that

$$\text{Prob}\{\sup_{K_1 \leq k \leq K_2} \|p(k) - p^0\| > \varepsilon\} = 0, \ \forall \alpha < \alpha_0 \qquad (21)$$

Thus, if the ODE has an asymptotically stable configuration $p^0$, then for all initial conditions, which are sufficiently close to $p^0$, the cGA essentially converges to $p^0$. ⋄

In the rest of the analysis, we consider the stability properties of the ODE and we talk in terms of stability, unstability, etc. about some configurations in $K$ and finally, we study the convergence of the cGA.

### 4.1 Stationary Configurations and the Stability Property

The following theorem characterizes the solutions of the ODE and hence, states the long-term behavior of the cGA.

**Theorem 2.** If the learning step is sufficiently small, the following is true for the cGA.

1. All deterministic configurations are stationary configurations.

2. All non-deterministic configurations are non-stationary configurations.

3. All local maximums of $g$ are asymptotically stable and the other points of $\Omega$ are unstable.

**Proof.**

1. By inspection of (15), if $p$ is a deterministic configuration, i.e. $p \in K^*$, then by Lemma 3 (5), $f(p) = 0$; therefore, $p$ is a stationary configuration.

2. Assume that $S = \{y | y \in \Omega, Pr(y|p) > 0\}$. It is clear that if $p$ is a non-deterministic configuration, then $N_S$, the cardinality of $S$, is an even number greater than one.

$$\sum_{i=1}^{n} |\tfrac{dp_i}{dt}| = \sum_{i=1}^{n} |f_i(p)| = \sum_{i=1}^{n} 2(\sum_{y} y_i Pr(y|p)| \sum_{g(z)<g(y)} Pr(z|p) - \sum_{g(z)>g(y)} Pr(z|p)|)$$
$$= 2 \sum_{y \in S} \{\Pr(y|p)| \sum_{g(z)<g(y)} Pr(z|p) - \sum_{g(z)>g(y)} Pr(z|p)|\}(\sum_{i=1}^{n} y_i) \quad (22)$$

$g$ is an injective function so we have

$$S = \{y^i |\ y^i \in \Omega\ and\ 1 \le i \le N_S\}$$
$$where\ N_S > 1\ and\ \forall i < j,\ g(y^i) > g(y^j) \quad (23)$$

according to (23), (22) can be rewritten as

$$\sum_{i=1}^{n} |\tfrac{dp_i}{dt}| = 2 \sum_{k=1}^{N_S} Pr(y^k|p)(\sum_{i=1}^{n} y_i)| \sum_{j=1}^{k-1} Pr(x^j|p) - \sum_{j=k+1}^{N_S} Pr(x^j|p)| \quad (24)$$

by inspection of (24) and taking into account that $N_S > 1$,

$$\sum_{i=1}^{n} |\tfrac{dp_i}{dt}| > 0$$

therefore, at least there is one $i$ where

$$\frac{dp_i}{dt} \neq 0$$

and consequently, $p$ is not a stationary configuration.

Note that if the function $g$ is not injective, the result may be invalid and we cannot make sure that all non-deterministic configurations are non-stationary configurations.

3. To prove this part of the theorem, we apply Lyaponov's indirect method (Drazin, 1992) to $f(p^0)$ where $p^0 \in K^*$ ($p^0$ can be considered as a binary string that belongs to $\Omega$). At first, we compute the Jacobian Matrix of $f(.)$ in $p^0$,

$$\left.\frac{\partial f_i(p)}{\partial p_m}\right|_{p^0} = 2\{\sum_y \{\overbrace{y_i \left.\frac{\partial Pr(y|p)}{\partial p_m}\right|_{p^0} (\sum_{g(z)<g(y)} Pr(z|p^0) - \sum_{g(z)>g(y)} Pr(z|p^0))}^{T_1^{i,m}(p^0)}\}$$
$$+ \sum_y \{\overbrace{y_i Pr(y|p^0)(\sum_{g(z)<g(y)} \left.\frac{\partial Pr(z|p)}{\partial p_m}\right|_{p^0} - \sum_{g(z)>g(y)} \left.\frac{\partial Pr(z|p)}{\partial p_m}\right|_{p^0})}^{T_2^{i,m}(p^0)}\}\} \quad (25)$$

We split $\Omega$ into three subspaces: $\Omega_1 = \{p^0\}$, $\Omega_2 = \{y|d_H(p^0,y) = 1\}$, and $\Omega_3 = \{y|d_H(p^0,y) \geq 2\}$. Assume that

$$w \in \Omega_2, \ w_m \neq p_m^0 \text{ and } \forall i \neq m, \ w_i = p_i^0 \quad (26)$$

By Lemma 3 and some simplification, two parts of (25) can be rewritten as

$$T_1^{i,m}(p^0) = w_i \left.\frac{\partial Pr(w|p)}{\partial p_m}\right|_{p^0} \{I(g(p^0) < g(w)) - I(g(p^0) > g(w))\} \quad (27)$$

and

$$T_2^{i,m}(p^0) = p_i^0 \left.\frac{\partial Pr(w|p)}{\partial p_m}\right|_{p^0} \{I(g(w) < g(p^0)) - I(g(w) > g(p^0))\} \quad (28)$$

by (27) and (28),

$$\left.\frac{\partial f_i(p)}{\partial p_m}\right|_{p^0} = 2 \left.\frac{\partial Pr(w|p)}{\partial p_m}\right|_{p^0} (w_i - p_i^0)\{I(g(p^0) < g(w)) - I(g(p^0) > g(w))\} \quad (29)$$

where the value of $I(exp)$ is one when $exp$ is true and it is zero when $exp$ is false. If $i \neq m$, by (26) we have

$$w_i = p_i^0 \text{ or } w_i - p_i^0 = 0 \quad (30)$$

therefore,

$$\left.\frac{\partial f_i(p)}{\partial p_m}\right|_{p^0} = 0 \tag{31}$$

For $i = m$, we investigate two cases: 1) $p^0$ is a local maximum 2) $p^0$ is not a local maximum. If $p^0$ is a local maximum, i.e. for each $y \in \Omega_2$, $g(p^0) \geq g(y)$, then by (29)

$$\left.\frac{\partial f_m(p)}{\partial p_m}\right|_{p^0} = -2 < 0 \tag{32}$$

and by (31) and (32),

$$J(f(p^0)) = Diag\{\overbrace{-2, ..., -2}^{n}\} \tag{33}$$

Thus, all eigenvalues of $J(f(p^0))$ are -2 and by Lyaponov's indirect method, $p^0$ is an asymptotically stable stationary configuration.

If $p^0$ is not a local maximum, then there is at least one $v \in \Omega_2$ and there exists an index $q$ where

$$g(v) > g(p^0), \ v_q \neq p_q^0 \tag{34}$$

In this case, for $i = m = q$, by (29) we write

$$\left.\frac{\partial f_q(p)}{\partial p_q}\right|_{p^0} = 2 > 0 \tag{35}$$

By (31) and (35), we conclude that $J(f(p^0))$ is a diagonal matrix, where at least one of its eigenvalues is positive and by Lyaponov's indirect method, $f(.)$ is unstable in $p^0$. ⋄

### 4.2 Convergence results

Based on Theorem 2, we can conclude that the cGA will never stay in a configuration, which is not a local maximum of $g$. This still leaves one question unanswered. Is it possible that $p(k)$ does not converge to a local maximum of $g$, for example, if the algorithm exhibits limit cyclic or chaotic behavior? Regarding this question, we provide the necessary condition for the cGA to converge to a local maximum of $g$. This is proven in Theorem 3 below.

**Theorem 3.** For the initial configuration $p(0) = (0.5, ..., 0.5)$, the cGA always converges to a local maximum of $g$.
**Proof.** The function $f(.)$ is continuous on $K$, therefore, there is a differentiable function $F$ where

$$F : \Re^n \to \Re$$

$$\forall 1 \leq i \leq n \ and \ all \ p \in K, \ \frac{\partial F}{\partial p_i}(p) = f_i(p) \tag{36}$$

Now, consider the variation of $F$ along the trajectories of the ODE. By (17) and (36),

$$\frac{dF}{dt}(p) = \sum_{i=1}^{n} \frac{\partial F}{\partial p_i}(p)\frac{dp_i}{dt} = \sum_{i=1}^{n} f_i(p)\frac{dp_i}{dt} = \sum_{i=1}^{n} (\frac{dp_i}{dt})^2 \geq 0 \qquad (37)$$

Thus, $F$ is non-decreasing along the trajectories of the ODE. Also, due to the nature of the algorithm given by (1), for the initial configuration $(0.5, ..., 0.5)$, the solution of the ODE (17), will be confined to $K$ which is a compact subset of $\Re^n$. Hence, by LaSalle's invariance principle ((Narendra and Annaswamy, 1989), Theorem 2.7), asymptotically, the trajectories will be in the set $K_1 = \{p \in K | (dF/dt)(p) = 0\}$. By (37),

$$\frac{dF}{dt}(p) = 0 \;\Rightarrow\; f_i(p) = (\frac{dp_i}{dt}) = 0 \;,\; 1 \leq i \leq n \qquad (38)$$

Therefore, $p$ is a stationary configuration of the ODE. Thus, the solution should converge to a stationary configuration. Since by Theorem 2 all stationary configurations which are not local maxima are unstable, the theorem follows. ⋄

Theorems 2 and 3 together characterize the long-term behavior of the cGA when the function $g$ is injective. Theorem 2 states that only local maxima of $g$ are asymptotically stable stationary configurations of the algorithm. In addition, Theorem 3 shows that the cGA cannot converge to any point in $K$ which is not a local maximum. If the function $g$ is not an injective function, then Theorem 2 (part 2) may be invalid and we cannot make sure that the local maxima of $g$ are the only stable stationary configurations of the cGA. In this case, the cGA may converge to some non-deterministic configurations and stay at them.

## 5 Conclusion

The cGA is an estimation of distribution algorithm. It is very simple and can be easily implemented in hardware. Using a small amount of memory, it can have many applications in the memory constraint problems. In this paper, a mathematical framework of the cGA, based on the weak convergence and the non-linear systems theories, has been proposed and consequently, the convergence behavior of the cGA has been studied. We have proven that the local maxima of an injective function are asymptotically stable stationary points of the cGA and shown that the cGA converges to one of these local maxima. While the results obtained in this paper are interesting by their own, they can also serve as one of the first steps towards using ODE in analysis of EDAs and the other evolutionary algorithms.

There are a lot of open questions and we have planned to study them in the future. First of all, we are interested in extending our framework to non-injective functions and determining the convergence rate of the cGA for different functions. Recently, some theorems have been developed in the stochastic approximation theory that can be useful in this regard (Kushner and Yin, 2000). Since for each optimization algorithm chosen to solve a problem, the shape and size of the basin of attractions may be different, we also would like to compute the basin of attractions of local maxima for a determined function. Comparing the basin of attractions of local maximum for the cGA with the basin of attractions of local maxima for other algorithms help us in choosing a better algorithm for optimization of a determined function. For example if we could show that the basin of attraction of global maximum for the cGA is bigger than the basin of attraction of the same point for the PBIL then we can predict that for different initial values the cGA will converge to the global maxima with a higher probability.

## References

Ahn, C. W. and Ramakrishna, R. S. (2003). Elitism-based compact genetic algorithms. *IEEE Trans. Evolutionary Computation*, 7(4):367–385.

Baluja, S. (1994). Population-based incremental learning: A method for integrating genetic search based function optimization and competitive learning. Technical Report CMU-CS-94-163, CMU, Pittsburgh, PA, USA.

Baluja, S. and Caruana, R. (1995). Removing the genetics from the standard genetic algorithm. In Prieditis, A. and Russel, S., editors, *In the Proceedings of the Int. Conf. on Machine Learning 1995*, pages 38–46, San Mateo, CA, USA. Morgan Kaufmann Publishers.

Baluja, S. and Davies, S. (1997). Using optimal dependency-trees for combinatorial optimization: Learning the structure of the search space. Technical Report CMU-CS-97-107, CMU, Pittsburgh, PA, USA.

Baraglia, R., Hidalgo, J. I., and Perego, R. (2001a). A hybrid heuristic for the traveling salesman problem. *IEEE Trans. Evolutionary Computation*, 5(6):613–622.

Baraglia, R., Perego, R., Hidalgo, J. I., Lanchares, J., and Tirado, F. (2001b). A parallel compact genetic algorithm for multi-fpga partitioning. In *the Proceedings of the Ninth Euromicro Workshop on Parallel and Distributed Processing*, pages 113–120, Mantova, Italy. IEEE Computer Society.

De Bonet, J. S., Isbell Jr., C. L., and Viola, P. A. (1996). Mimic: Finding optima by estimating probability densities. In *the Proceedings of NIPS96*, pages 424–430, Denver, CO, USA. MIT Press.

Drazin, P. G. (1992). *Nonlinear Systems*. Cambridge University Press.

Droste, S. (2005). Not all linear functions are equally difficult for the compact genetic algorithm. In *the Proceedings of GECCO 2005*, pages 679–686, Washington DC, USA.

Gallagher, J. C., Vigraham, S., and Kramer, G. R. (2004). A family of compact genetic algorithms for intrinsic evolvable hardware. *IEEE Trans. Evolutionary Computation*, 8(2):111–126.

Goldberg, D. E. (1989). *Genetic Algorithms in Search, Optimization, and Machine Learning*. Reading, MA: Addison-Wesley.

Gonzalez, C., Lozano, J. A., and Larranaga, P. (2000). Analyzing the pbil algorithm by means of discrete dynamical systems. *Complex Systems*, 12(4):465–479.

Harik, G. R., Lobo, F. G., and Goldberg, D. E. (1999). The compact genetic algorithm. *IEEE Trans. Evolutionary Computation*, 3(4):287–297.

Hohfeld, M. and Rudolph, G. (1997). Towards a theory of population based incremental learning. In *the Proceedings of the 4th IEEE Conferences on Evolutionary Computation*, pages 1–5, Indianapolis, IN, USA. IEEE Press.

Kushner, H. J. (1984). *Approximation and Weak Convergence Methods for Random Processes*. Cambridge, MA: MIT Press.

Kushner, H. J. and Yin, G. (2000). *Stochastic Approximation and Recursive Algorithms and Applications*. Springer Verlag.

Larranaga, P. and Lozano, J. A. (2001). *Estimation of Distribution Algorithms: A New tools for Evolutionary Computation*. Kluwer Academic Publishers.

Muhlenbein, H. (1998). The equation for response to selection and its use for prediction. *Evolutionary Computation*, 5(3):303–346.

Muhlenbein, H. and Mahnig, T. (1999a). Convergence theory and application of the factorized distribution algorithm. *Journal of Computing and Information Technology*, 7(1):19–32.

Muhlenbein, H. and Mahnig, T. (1999b). The factorized distribution algorithm for additively decomposed functions. In *the Proceedings of the Congress on Evolutionary Computation 1999*, pages 752–759, Washington DC, USA.

Narendra, K. S. and Annaswamy, A. (1989). *Stable Adaptive Systems*. Englewood Cliffs: Printice Hall.

Neil, J. R. and Rowe, J. E. (2005). Stable fixed points of fixed-graph estimation of distribution algorithms in the infinite population limit. Technical report, The University of Birmingham, UK.

Pelikan, M., Goldberg, D. E., and Lobo, F. (1999). A survey of optimization by building and using probabilistic model. Technical Report 99018, Illinois University, Illinois, USA.

Rastegar, R. and Meybodi, M. R. (2005a). A new estimation of distribution algorithm based on learning automata. In *the Proceedings of IEEE Conference on Evolutionary Computation 2005*, pages 1982–1986, UK.

Rastegar, R. and Meybodi, M. R. (2005b). A note on the population based incremental learning with infinite population size. In *the Proceedings of IEEE Conference on Evolutionary Computation 2005*, pages 198–205, UK.

Rastegar, R. and Meybodi, M. R. (2005c). A study of the global convergence time complexity ofestimation of distribution algorithms. In *Lecture Notes in Artificial Intelligence*, volume 3641, pages 441–450, Regina, Canada. Springer.

Sastry, K. and Goldberg, D. E. (2000). On extended compact genetic algorithm. In *the Proceedings of GECCO 2000*, pages 352–359, San Francisco, CA, USA.

Syswerda, G. (1992). Simulated crossover in genetic algorithm. In *the Proceedings of FOGA-2*, pages 239–255, San Mateo, CA, USA. Morgan Kaufmann Publishers.

Zhang, Q. (2004a). On stability of fixed points of limit models of univariate marginal distribution algorithm and factorized distribution algorithm. *IEEE Trans. Evolutionary Computation*, 8(1):80–93.

Zhang, Q. (2004b). On the convergence of a factorized distribution algorithm with truncation selection. *Complexity*, 9(4):17–23.

Zhang, Q. and Muhlenbein, H. (2004). On the convergence of a class of estimation of distribution algorithms. *IEEE Trans. Evolutionary Computation*, 8(2):127–136.